# Wearable multi-color RAPD screening device

Arda Gulersoy, Ahmet Berk Tuzcu, Doga Gunduzalp, Koray Kavakli, Abdullah Kucukoduk, et al.

# Wearable Multi-color RAPD Screening Device

Arda Gulersoy*[a], Ahmet Berk Tuzcu*[a], Doga Gunduzalp*[a], Koray Kavakli*[a],
Abdullah Kucukoduk*[a], Umit Yasar Guleser*[b], Ugur Aygun*[a], Murat Hasanreisoglu*[b],
Afsun Sahin*[b], Hakan Urey*[a]

[a]Dep. of Electrical and Electronics Engineering, Koc University, Istanbul, Turkey; [b]School of Medicine, Koc University, Istanbul, Turkey

**ABSTRACT**

In this work, we developed a wearable, head-mounted device that automatically calculates the precise Relative Afferent Pupillary Defect (RAPD) value of a patient. The device consists of two RGB LEDs, two infrared cameras, and one microcontroller. In the RAPD test, the parameters like LED on-off durations, brightness level, and color of the light can be controlled by the user. Upon data acquisition, a computational unit processes the data, calculates the RAPD score and visualizes the test results with a user-friendly interface.Multiprocessing methods used on GUI to optimize the processing pipeline. We have shown that our head-worn instrument is easy to use, fast, and suitable for early-diagnostics and screening purposes for various neurological conditions such as RAPD, glaucoma, asymmetric glaucoma, and anisocoria.

## 1. INTRODUCTION

Pupil light reflex is a reflex which causes change in the pupil diameter, due to a bilateral or unilateral light illumination. When light is flashed into one or both eyes, for a healthy person, the PLR arcade causes both pupils to contract simultaneously and eanlarge after the light is removed. RAPD(relative afferent pupillary defect), which is the problem linked with the pupil dilation rate of a person, causes people to dilate on or two of his eyes with a different reaction due to a spoiling on one of the afferent visual pathways.With the various of studies it is proven that there is a clear connection with the PLR and retinal and optic nerve related disorders. Furthermore early detection is a key element to prevent diseases to turn into blindness and according to the World Health Organization(WHO) 1 billion of 2.2 billion eye disorders can be prevented with early detection.Also as Tham states there is an unequal prevalence of glaucoma, is a PLR pattern related disease which 111 million people suffer by the year of 2040 , across the globe, emphasizing the need for cost-effective and timely diagnostic techniques  The Swinging Flashlight Test (SFT) is a technique used to identify RAPD , which means comparing the response of both eyes to light. In this test, a medical staff shines a light into the patient's eye and attempts to monitor changes in pupil size due to light stimulation.However,  SFT, which is manual, relies heavily on the human factor and it makes hard to identify low RAPD cases , also requires trained staff .Furthermore, the duration of the light stimulus is uncontrolled and can vary, makes standardization of the test difficult. Conventional non-invasive methods cannot assess the patient's chromatic RAPD score, which is important for glaucoma detection [1].In this research, we developed a low-cost head-mounted RAPD instrument that can be used in low-resource environments without trained personnel.RAPD measurement device , as shown in Figure 1., consists of two infrared cameras, two RGB LEDs, and a frame-mounted microcontroller(Arduino Uno).With the help of headset , it is possible to adjust the camera according to  patient's interpupillary distance. Also the parameters of the test such as  LED on-off durations and brightness levels can be adjusted via GUI, just before the test.Moreover, the user can set the color of the LEDs with options red, blue, green, and white.Chromatic assessment of the RAPD important because of the cones photoreceptors that inhabit in the brain.They are highly sensitive to the chromatic light, so that can build a connection between the wavelength of the chromatic light and neural pathway disorders with the help of investigation of PLR.Our device can be also used in that direction.

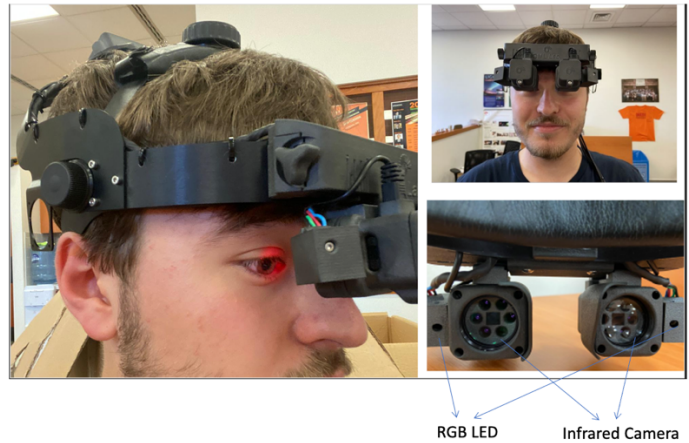

Figure 1: Front, back and side view of RAPD measurement device

## 2.METHOD

Pupil diameter (D) is calculated for each frame, to plot the change of the pupil diameters with respect to time. RAPD is calculated by the logarithm of the median of the magnitude, which means the percentage change from the rest state to the constriction state shown in Fig 3a, of the left eye divided by the median of the magnitude of the left eye. In short, RAPD can be calculated using the following equations;

In detail base of our device is composed of a 3D-printed frame.Illumination is powered by an Arduino Uno micontroller and the fine adjustment related with the illumination such as, LED on/off time, LED power can be controlled via user interface.For the pupil tracker, eye is illuminated by six 940nm IR LED and visible light is blocked with an IR filter. To enhance the output of the high resolution camera, a bi-convex lens (Thorlabs LB1471) added infront of the both cameras.

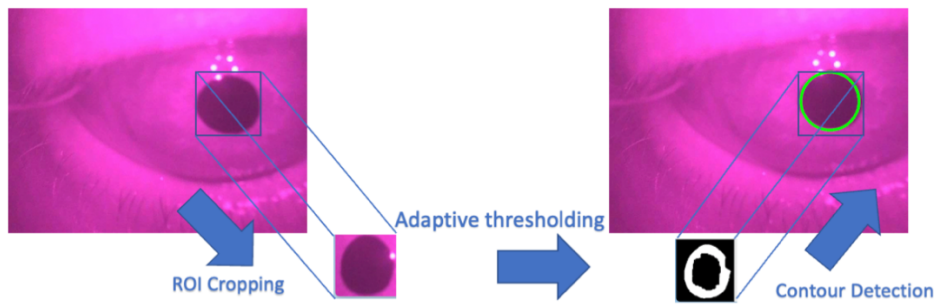

Figure 2: Pipeline of the eye tracking algorithm

Captured infrared images are processed to detect pupils as shown in Fig 2.The programming language that is used to process frame is Python and the main library is OpenCV. First, we apply convolutional Gaussian blur with the kernel size 7 to the captured frames to reduce the noise .Then we find the region of interest, which is the pupil ,by applying a nonadaptive threshold. Later from the binary image we find the blackest region and crop the frame relative to that to obtain the region of interest(ROI).After, an adaptive threshold applied to the region of interest(ROI) of the original infrared frame to obtain a binary frame. Finally, we find contour coordinates from the binary frame to draw the best fitting circle around the pupils to track the pupil diameter during the test, which can be seen in Fig 2 .Blinking and other noise related signals are handled via post processing methods such as low pass filtering and interpolation.

For each frame a pupil diameter(D) is calculated and stored into a array, to illustrate the change of the diameters during the test with respect to time later. Since the synchronization of the both the trackers and LED controller is important to calculate the latency, a multithreading approach is used to run each tracker, LED controller and GUI sections on different cores of the processors to boost the speed of the pipeline.

RAPD is calculated by the logarithm of the median of the magnitude, which means the percentage change from the rest state to the constriction state shown in Fig 3a, of the left eye divided by the median of the magnitude of the left eye. In short, RAPD can be calculated using the following equations;

$$Magnitude(\%) = \frac{D_{rest} - D_{constricted}}{D_{rest}} * 100 \quad (1)$$

$$RAPD = 10 * \log_{10} \frac{median(Magnitude_{right})}{median(Magnitude_{left})} \quad (2)$$

where *D_rest* and *D_constricted* are the pupil diameters in rest and constricted states, respectively. For the classification of the patients, we classify a patient who has RAPD score lower than 0.5 as negative and greater than 1 as RAPD positive patient. Since we are seeking for relative changes, there is no need for precise pupil diameter calculation.

## 3.SOFTWARE IMPLEMENTATION

To enhance the availability of the device a GUI was designed so that the user can control all the parameters and observe the results.As can be seen oni Fig 3c , with the opening page of the interface, user can save the information of the patient that the test will be conducted on , such as age, sex, medical history , etc.Also parameters of the test is being saved for later usage.After that, a page on Fig 3a that shows the preview of the infrared camera that helps align the position of the eyes for best performance, this page can be also activated during the test using multi-threading computing. The user interface uses a multi-core CPU and four different threads. The first of these threads manage interface operations. Two of these threads are used to continuously stream the right and left eyes in real-time. The fourth thread ensures that the desired LEDs flash via the Arduino when the test starts, and the images of the right and left eyes and the relevant times are recorded in the desired time stamps. In this way, the test can be implemented ,the desired data can be recorded, and the eye stream can be examined simultaneously..Lastly when the test is finished a result page pops up that shows the all variables that calculated on test, also with the plot that contains information about LED on/off times and pupil diameters during the test.

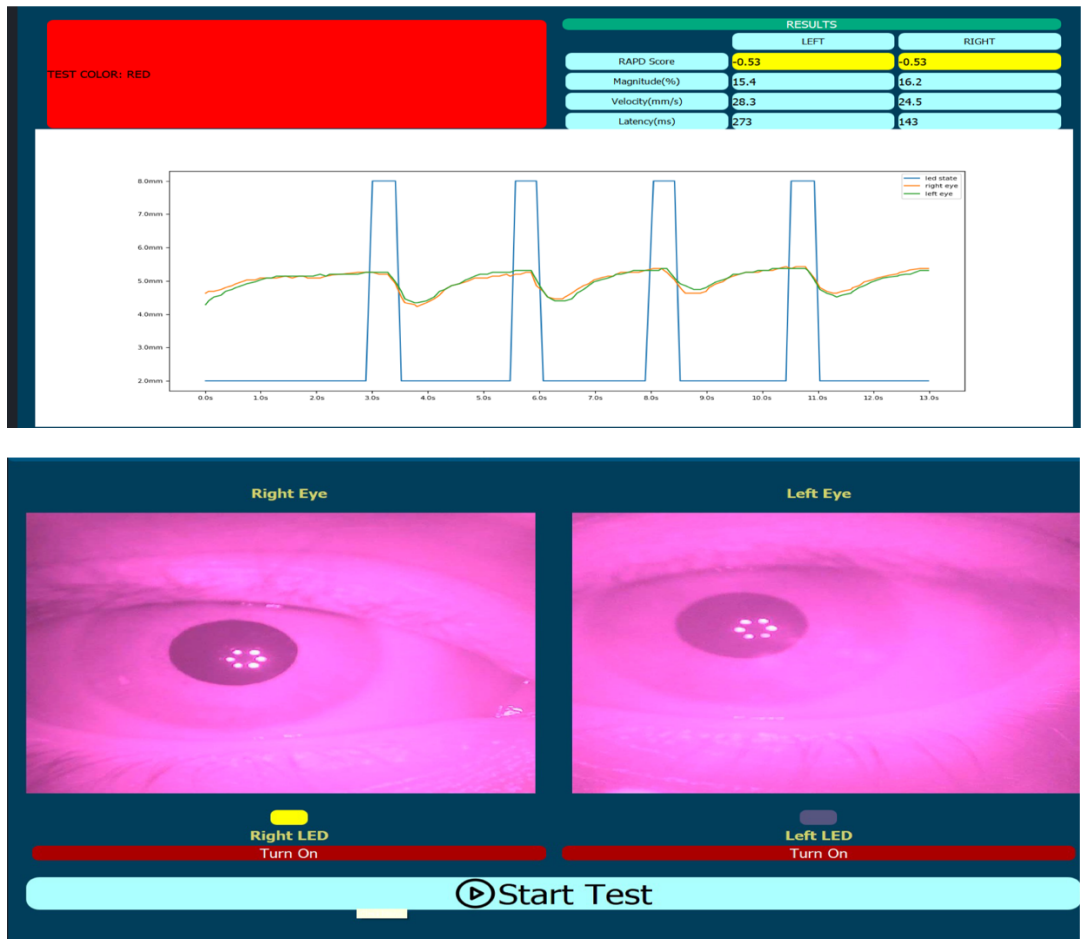

Fig3 . Preview and results page of GUI

## 4.SUMMARY

The pupil light reflex (PLR) is a reflex that causes changes in pupil size in response to light simulation in one or both eyes. A relative afferent pupillary defect(RAPD) occurs when the pupil dilation rate is different in one or both eyes due to damage to an afferent visual pathway.The Swinging Flashlight Test (SFT) is a manual method used to identify RAPD, but it is not always accurate. In this research we successfully developed a low cost head mounted RAPD instrument that can be used in low resource settings without trained personnel.The instrument allows for the adjustment of LED parameters and colors, also it can be used for chromatic assessment of RAPD.

**Acknowledgments**

The authors would like to thank Selim Olçer, Fırat Turkkal, Gunes Aydindogan and Cem Kesim for their help in prototype development.  This work has been supported European Innovation Council (EIC) Horizon under grant agreement no 101057672.